\def\year{2022}\relax
\documentclass[letterpaper]{article} 
\usepackage{aaai22}  
\usepackage{times}  
\usepackage{helvet}  
\usepackage{courier}  
\usepackage[hyphens]{url}  
\usepackage{graphicx} 
\usepackage{natbib}  
\usepackage{caption} 
\DeclareCaptionStyle{ruled}{labelfont=normalfont,labelsep=colon,strut=off} 
\usepackage{algorithm}
\usepackage{algorithmic}

\usepackage{multirow}
\usepackage{subfigure}
\usepackage{booktabs}
\usepackage{amsmath}
\usepackage{makecell}
\usepackage{xcolor}

\usepackage{newfloat}
\usepackage{listings}
\floatstyle{ruled}
\newfloat{listing}{tb}{lst}{}
\floatname{listing}{Listing}

\title{Diaformer: Automatic Diagnosis via Symptoms Sequence Generation}

\author{
    Junying Chen\textsuperscript{\rm 1}\equalcontrib,
    Dongfang Li\textsuperscript{\rm 1}\equalcontrib,
    Qingcai Chen\textsuperscript{\rm 1,2}\thanks{Corresponding author.},
    Wenxiu Zhou\textsuperscript{\rm 1},
    Xin Liu\textsuperscript{\rm 2}
}
\affiliations{

    \textsuperscript{\rm 1} Harbin Institute of Technology (Shenzhen)\\
    \textsuperscript{\rm 2} Peng Cheng Laboratory\\
    \{junying.chen.cs, crazyofapple, hit.liuxin\}@gmail.com, qingcai.chen@hit.edu.cn, wen.xiu.zhou@outlook.com 

}

\usepackage{bibentry}

\begin{document}

\maketitle

\begin{abstract}
Automatic diagnosis has attracted increasing attention but remains challenging due to multi-step reasoning. Recent works usually address it by reinforcement learning methods. However, these methods show low efficiency and require  task-specific reward functions. Considering the conversation between doctor and patient allows doctors to probe for symptoms and make diagnoses, the diagnosis process can be naturally seen as the generation of a sequence including symptoms and diagnoses. Inspired by this, we reformulate automatic diagnosis as a symptoms Sequence Generation (SG) task and propose a simple but effective automatic \textit{Dia}gnosis model based on Trans\textit{former} (Diaformer). We firstly design the symptom attention framework to learn the generation of symptom inquiry and the disease diagnosis. To alleviate the discrepancy between sequential generation and disorder of implicit symptoms, we further design three orderless training mechanisms. Experiments on three public datasets show that our model outperforms baselines on disease diagnosis by 1\%, 6\% and 11.5\% with the highest training efficiency. Detailed analysis on symptom inquiry prediction demonstrates that the potential of applying symptoms sequence generation for automatic diagnosis.
\end{abstract}
\section{Introduction}

Automatic diagnosis has recently attracted increasing attention from researchers because of its potential in simplifying diagnostic procedures~\cite{tang2016inquire,kao2018context}, helping make better and more effective diagnostic decisions~\cite{shivade2014review,xia2020generative}, and even helping build a diagnostic dialogue system as a dialogue management~\cite{li2017end,wei2018task,xu2019end,Teixeira2021TheIO}. An automatic diagnosis system is built on conversations between the agent and the patient where allows the agent to probe for symptoms and make diagnoses. As an example is shown in Table~\ref{EHR}, the agent interacts with users to inquiry about additional symptoms (i.e., implicit symptoms) beyond their self-reports (i.e., explicit symptoms) and make a disease diagnosis at the end. When inquiring about additional symptoms, the automatic diagnosis system can only obtain the value of symptoms in the ``implicit\_symptoms'' set or get a ``not\_sure'' answer for outside symptoms. 
\begin{table}[t]
  \centering
  \begin{tabular}{p{8cm}}
\toprule
\textbf{explicit\_symptoms} (the symptoms from self-reports):\\
\{cough:\emph{true},  snot:\emph{true}\} \\ \midrule
\textbf{implicit\_symptoms} (the symptoms from conversation): \\ 
\{sore throat:\emph{true}, fever:\emph{true}, harsh respiration:\emph{false}\} \\ \midrule
\textbf{disease\_tag} (the target disease):\\
bronchitis of childhood \\ \bottomrule
\end{tabular}
\caption{An example of automatic diagnosis data.}
\label{EHR}
\end{table}
Thus, the disease diagnosis task can be defined as inquiring the implicit symptoms step by step with limited interaction turns, and then diagnosing the disease based on explicit symptoms and the additional symptoms inquired. Note that different from the dialogue system of automatic diagnosis, automatic diagnosis we called here is symptom checking task in \cite{tang2016inquire}, which also serves as dialogue manager in task-oriented dialogue system of automatic diagnosis~\cite{wei2018task,xu2019end,add3,xia2020generative,Teixeira2021TheIO}.

Due to the existing of implicit symptoms, this task can be considered as a multi-step reasoning problem. The challenge of the task is how to capture the underlying dynamics and uncertainties of reasoning process, then inquiry about accurate symptoms under small labeled data and limited turns. Most previous methods usually address this problem as a sequential decision-making process, then formulate the process as Markov Decision Processes (MDPs) and employ Reinforcement Learning (RL) for policy learning~\cite{tang2016inquire,DBLP:journals/corr/PengLLGCLW17,DBLP:journals/corr/abs-1711-10712,add2,kao2018context,wei2018task,xu2019end,liao2020task,xia2020generative,Hou2021ImperfectAD,Teixeira2021TheIO}. However, RL learns how to inquire about symptoms and make a disease diagnosis only with final accuracy rewards, which partly deviates from the actual doctor’s diagnostic process. In real clinical diagnosis scenario, doctors carefully select relevant questions and ask patients with a medical diagnostic logic~\cite{xia2020generative}. 
The policy learning of RL tries to learn which symptom inquiry improves the rewards but not the doctor’s diagnostic logic directly. As a result, RL relies on the random trials to learn how to improve the reward, but don't learn directly the correlation among symptoms and the standard diagnosis paradigm. It leads to low efficiency in learning how to make symptom inquiry decisions. 
Besides, there is still no explicit solution to find ideal reward functions, which may make the RL-based model hard to balance the decision learning between disease diagnosis and symptom inquiry.

Considering the diagnosis process can be naturally seen as the generation of a sequence, we reformulate automatic diagnosis as a Sequence Generation (SG) task in this work. Different from RL-based methods, the multi-step inquiry process is explicitly modeled at generating a sequence including symptoms and diagnoses. This can improve the efficiency and explainability of multi-step reasoning process. Moreover, the latent relationship among previous explicit symptoms and current symptom can be learned. Hence, the accurate inquiry of implicit symptoms would help to improve the accuracy of disease diagnosis, which is similar to doctors' diagnostic logic.
As the example shown in Table~\ref{EHR}, RL-based models tentatively learn which symptom inquiry helps to predict the \emph{children’s bronchitis} target using policy learning in large state/action spaces. By contrast, the SG-based model learn to inquire \emph{sore throat}, \emph{fever} and \emph{brash breath} sequentially based on the explicit symptoms, so that the model can learn the latent relationship of symptoms inquiry and the diagnosis decision-making more efficiently. 

As a step forward, we propose a simple but effective  automatic \textit{Dia}gnosis model based on Trans\textit{former} (Diaformer).
It consists of a symptom attention framework and learns with three orderless training mechanisms.
The symptom attention framework is introduced to model the automatic diagnosis using the Transformer architecture~\cite{vaswani2017attention}. The self-attention mechanism of Transformer can reduce the position dependence and learn multiple relationships among the multiple symptoms. To consider previous symptoms in current step, we also propose a attention mask mechanism. 
Each implicit symptom can attend to the given explicit symptoms and the previous implicit symptoms, while each explicit symptom can only see the explicit symptoms. 
As we learn the symptom inquiry by symptoms sequence generation, there is a bias caused by the discrepancy between the order of symptoms sequence learned and the disorder of golden implicit symptoms. To address this challenge, we further propose three orderless training mechanisms: sequence shuffle, synchronous learning, and repeated sequence. The main idea is to encourage the model to inquire symptoms in an orderless but accurate way, whereby improving the generalizability at inference time. Extensive experiments on MuZhi dataset~\cite{wei2018task}, Dxy dataset~\cite{xu2019end} and Synthetic dataset~\cite{liao2020task} show that our proposed model (Diaformer) outperforms baselines on disease diagnosis by 1\%, 6\% and 11.5\% with the highest training efficiency. Further analysis on symptom inquiry prediction demonstrates applying symptoms sequence generation is an plausible way to automatic diagnosis task.

Our contributions are summarized as follows: 
\begin{itemize}
\item To the best of our knowledge, we are the first to apply symptoms sequence generation for automatic diagnosis. We further show that our method can be applied under few conversation turns scenarios.  
\item We propose three orderless training mechanisms to alleviate the discrepancy between the sequential generation and the disorder of given implicit symptoms sets. The ablation studies show that these orderless mechanisms can significantly alleviate this bias.
\end{itemize}

\section{Preliminaries}
\begin{figure*}[ht]
  \centering
  \includegraphics[width=1\textwidth]{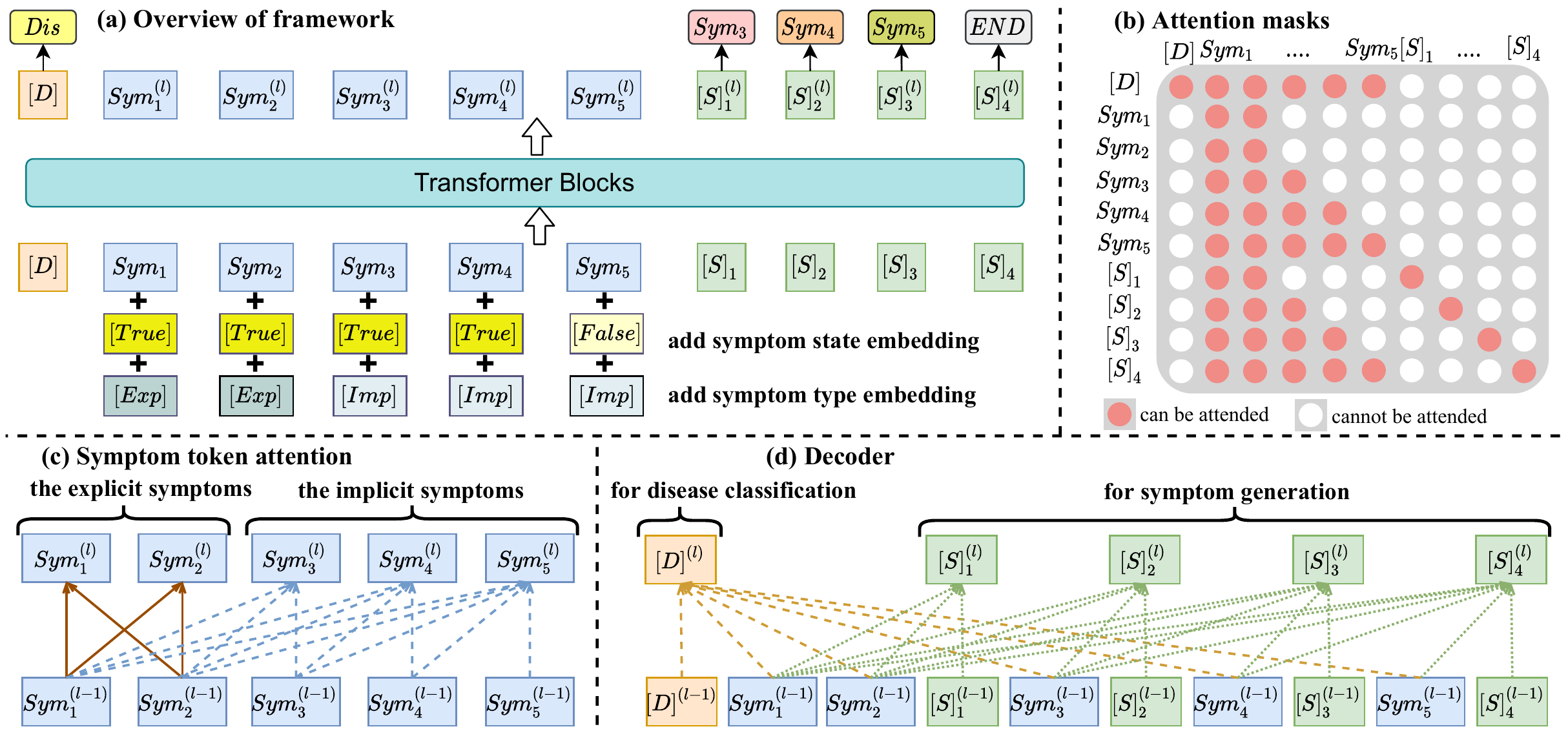}
  \caption{\label{overview}Illustration of symptom attention framework.}
\end{figure*}

In this section, we first formulate the automatic diagnosis task as a sequence generation (SG) task. Formally, an automatic diagnosis data includes an explicit symptom set $S_{exp}$, an implicit symptom set $S_{imp}$ and a target disease $Dis$.
As shown in Figure~\ref{overview}, $S_{exp}=\{Sym_{1},Sym_{2}\}$ and $S_{imp}=\{Sym_{3},Sym_{4},Sym_{5}\}$. For the task, the automatic diagnosis system can only access the explicit symptoms $S_{exp}$ at the beginning. Then the system can inquire symptoms in limited turns to obtain the implicit symptoms in $S_{imp}$. When the symptom inquires a symptom, the user simulator will take one of the three answer including \emph{True} for the positive symptom, \emph{False} for the negative symptom, and \emph{Not sure} for the symptom that is not mentioned in user goal $S_{exp}\cup S_{imp}$. We denote $S_{add}$ which $S_{add}\subseteq S_{imp}$ as the additional symptoms that had been inquired by the system. In the end, the system is asked to make a disease diagnosis based on the explicit symptoms and the addition symptoms. The task objective of learning is maximize the likelihood of disease diagnosis $P(Dis\mid S_{exp}\cup S_{add})P(S_{add}\mid S_{exp})$. Since the implicit symptoms contain important information that inquired by the doctors, the intuition is that the more implicit symptoms the model inquires, the higher diagnosis accuracy model can get. We transfer task learning objective to maximize $P(S_{imp} \mid S_{exp})$ as well as:
\begin{equation}
\prod_{S_{add}\subseteq S_{imp}}\prod_{Sym\in S_{imp} - S_{add}}P(Sym \mid S_{exp},S_{add}) \label{ground_target}
\end{equation}
then we use a symptom attention framework, shown in Figure~\ref{overview}, to learn predicting the implicit symptoms sequentially. Thus, we need to change the orderless $S_{imp}$ into the ordered symptoms sequence $T_{imp}$ as the target generation sequence. According to the autoregressive generation, the probability of output token depends on all previous. The objective of model learning is transferred to maximize the likelihood of $T_{imp}$ generation:
\begin{equation}
\prod_{i=1}^{|T_{imp}|}P(T_{imp}^{i}\mid S_{exp}, T_{imp}^{<i}) \label{pred_target}
\end{equation}
where the $T_{imp}^{i}$ denote i-th symptom in $T_{imp}$ and $T_{imp}^{<i}$ denote all the symptoms in front of $T_{imp}^{i}$. Hence we change the symptom inquire task to a sequence generation task. Since the discrepancy between the order of $T_{imp}$ and the disorder of $S_{imp}$, the SG training objective Eq.(\ref{pred_target}) is unequal to the Eq.(\ref{ground_target}), which seriously hinder the performance in automatic diagnosis. We propose three training mechanisms to make training objective Eq.(\ref{pred_target}) approximate to  Eq.(\ref{ground_target}). 

\section{Methodology}
In this section, we introduce the Diaformer, which consists of symptom attention framework and the orderless training mechanisms. Then we elaborate the generative inference.

\subsection{Symptom Attention Framework}
The illustration of symptom attention framework is shown in Figure~\ref{overview}. In this framework, we adopt multiple stacked Transformer blocks to model the automatic diagnosis by SG. Each block contains a feed forward network and a multi-head attention, in which all input tokens share the parameters via self-attention~\cite{vaswani2017attention}.

\paragraph{Input Representation} As shown in Figure~\ref{overview}a, all the symptoms in $S_{exp}$ and $S_{imp}$ are converted to the specific token embedding. Different from Transformer, we remove the position embedding, and the symptom input representation is computed by summing the corresponding token embedding, symptom state embedding and symptom type embedding. For the symptom state embedding, $[True]$ and $[False]$ indicate positive symptom and negative symptom. For the symptom type embedding, $[Exp]$ and $[Imp]$ signal whether the symptom belong to $S_{exp}$ or $S_{imp}$. In addition, we add two special tokens [S] and [D], which are used to predict symptom and disease respectively.

\paragraph{Attention Masks} Figure~\ref{overview}b show the attention mask matrix M, which determines whether query and key can attend to each other in self-attention by modifying the attention weight $W=softmax(\frac{QK^T}{\sqrt{d_k}}+M)$~\cite{vaswani2017attention}. Specifically, M is assigned as:
\begin{equation}
    \label{attnweight}
    M_{ij} = \left\{ \begin{array}{ll}
0, & \textrm{can be attended}\\
-\infty, & \textrm{cannot be attended}
\end{array} \right.
\end{equation} 
, where $0$ and $-\infty$ indicate red point and white point in Figure~\ref{overview}b. With it, we can prevent symptom prediction from seeing leaked information in self-attention layer to achieve autoregressive generation training of implicit symptoms.

\paragraph{Symptom Token Attention} Before input the framework, we initialize a implicit symptoms sequence $T_{imp}$ by $S_{imp}$.
As shown in Figure~\ref{overview}c, in each multi-head attentions of Transformer block, each explicit symptom can merely see the explicit symptoms and each implicit symptom can see the explicit symptoms and the previous implicit symptoms. To be specific, the representations of symptom tokens are update in multi-head attention~\cite{vaswani2017attention} as:
\begin{equation}
    \label{encoder}
    \begin{split}
	 S_{exp}^{(l)} \gets \textrm{MH-Attn}(Q&=S_{exp}^{(l-1)},KV=S_{exp}^{(l-1)}) \\
	 T_{imp}^{i(l)} \gets \textrm{MH-Attn}(Q&=T_{imp}^{i(l-1)},\\
	KV&=[S_{exp}^{(l-1)},T_{imp}^{\le i(l-1)}])
\end{split}
\end{equation}
where $Q$, $K$, $V$ denote the query, key and value in multi-head attention, $[.]$ denotes concatenation along the sequence dimension, $T_{imp}^{i(l)}$ indicates the $l$-th Transformer block layer output of the i-th implicit symptom in $T_{imp}$ and $S_{exp}^{(l)}$ denotes $l$-th layer output of the explicit symptoms.

\begin{figure*}[ht]
  \centering
  \includegraphics[width=0.72\textwidth]{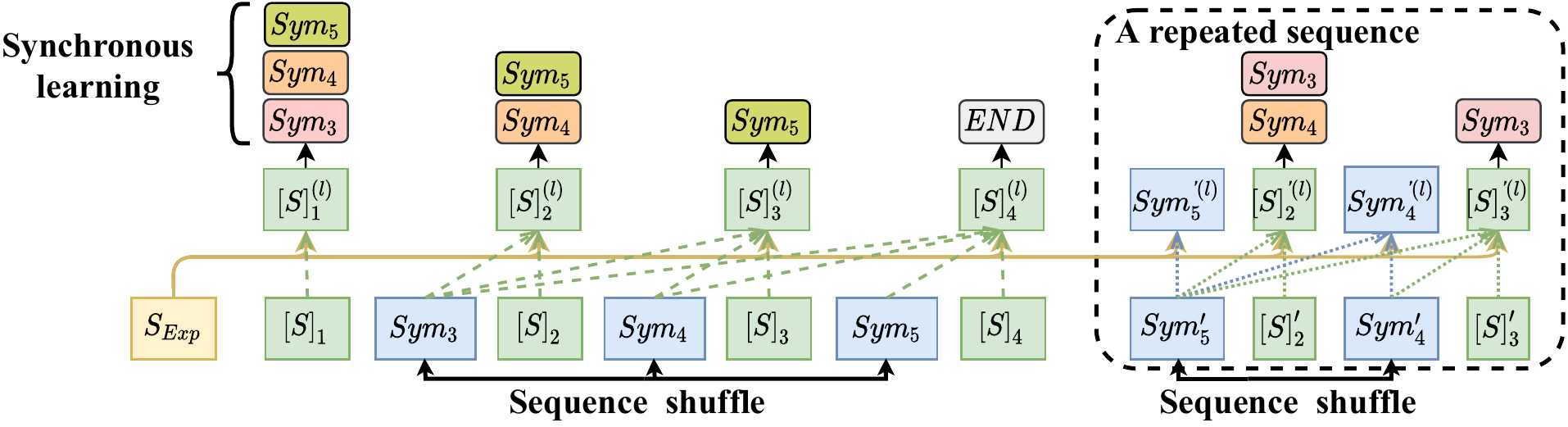}
  \caption{\label{orderless}Demonstrations of three orderless training mechanisms.}
\end{figure*}

\paragraph{Decoder}  
As shown in Figure~\ref{overview}a, we insert a $[S]$ sequence which length is (n+1), where n is the number of implicit symptoms. We used the $[S]$ sequence to learn generating the implicit symptoms as sequential symptom inquiry in the automatic diagnosis. The i-th $[S]$ in $[S]$ sequence is set to predict the i-th symptom in $T_{imp}$. And the last $[S]$ in the $[S]$ sequence is used to predict the action of ending inquiry, which is a \emph{END} symbol in Figure~\ref{overview}a. The attention flow of $[S]$ sequence is shown in Figure~\ref{overview}d. In the autoregressive symptoms sequence generation, each $[S]$ cannot see the target symptom and the implicit symptoms after the target symptom in $T_{imp}$, aimed to learn the target symptom without information leakage. Specifically, the representation of $[S]$ token is update as:
\begin{equation}
\label{sdecoder}
\begin{split}
[S]_{i}^{(l)} \gets \textrm{MH-Attn}(Q&=[S]_{i}^{(l-1)},\\KV&=[S_{exp}^{(l-1)},T_{imp}^{<i(l-1)},[S]_{i}^{(l-1)}])
\end{split}
\end{equation}
where $[S]_{i}^{(l)}$ denotes the l-th layer output of the i-th token in $[S]$ sequence. For the last layer outputs of $[S]$, we use a symptom classification layer that contains a liner layer weights $W_{sym}\in \mathcal{R}^{H\times C_{inq}}$, where H is the hidden size and $C_{inq}$ is the number of symptom inquiry types. For the symptom classification layer output $z$, we compute the cross entropy loss with softmax, i.e. $-log(softmax(z))$, as the symptom inquiry loss $\mathcal{L}_{sym}$.  As for the disease classification, we insert a special token $[D]$ to input sequence, shown in Figure~\ref{overview}a. As shown in Figure~\ref{overview}d, the token $[D]$ can attend all the symptoms. The vector of $[D]$ is updated as:
\begin{equation}
\label{ddecoder}
\begin{split}
[D]^{(l)} \gets \textrm{MH-Attn}(Q&=[D]^{(l-1)},\\KV&=[S_{exp}^{(l-1)},T_{imp}^{(l-1)},[D]^{(l-1)}])
\end{split}
\end{equation}
Similar to $[S]$, on the last layer output of $[D]$, we adopt a disease classification layer weights $W_{dis}\in \mathcal{R}^{H\times C_{dis}}$, where $C_{dis}$ is the number of disease types. We compute the cross entropy loss of it as the disease classification loss $\mathcal{L}_{dis}$. The final training loss is computed as $\mathcal{L} = \mathcal{L}_{dis}+\mathcal{L}_{sym}$.

\subsection{Orderless Training Mechanisms}
To alleviate the bias caused by the discrepancy between the order of the sequential generation of $T_{imp}$ and the disorder of implicit symptoms $S_{imp}$, We propose three orderless training mechanisms of SG, which is demonstrated in Figure~\ref{orderless} and detailed as follow.

\paragraph{Sequence Shuffle} We randomly shuffle the implicit symptoms sequence $T_{imp}$ to obtain new implicit symptoms sequence $T_{imp}^{'}$ in different order before input the model in each training step, so that the model can learning different order of symptom inquiry after multiple training epoch. It can prevent the model to impose over-fitting on inquiring symptoms in a specific order and fail to inquire the correct implicit symptom in a slightly different context. With the increase of training epoch, the model will gradually fit the symptom disorder distribution, whereby the $T_{imp}$ sequence generation approximate to the $S_{imp}$ inference. 

\paragraph{Synchronous Learning} While the model predicts the next symptom inquiry, we expect all the implicit symptoms, which have not been inquired, have equal probability to be inquired. As shown in Figure~\ref{overview}a, the $[S]_1$ is trained to predict the $Sym_3$. However, $Sym_3$, $Sym_4$ and $Sym_5$ should have the same priority to be inquired in the orderless set of the implicit symptoms. Therefore, we design the synchronous learning objective to train the model to synchronously predict the rest implicit symptoms that it can't see. As shown in the Figure~\ref{orderless}, each symptom prediction token $[S]$ is trained to predict all the rest implicit symptoms synchronously. Therefore, we use a concurrent softmax~\cite{peng2020large} replacing the original softmax to train $[S]$ to predict multiple symptoms synchronously. We remove the concurrent rate in ~\cite{peng2020large} and only use the concurrent softmax as a training mechanism, as we still use softmax in inference. The concurrent softmax can enable the model to learn multiple symptoms synchronously and eliminate the discrepancy between training and inference.
The reset implicit symptoms $S_{imp}-S_{add}$ is set as the training objective, so the concurrent softmax label $y$ is defined as:
\begin{equation}
    \label{labely}
    y_{i} = \left\{ \begin{array}{ll}
1, & y_{i}\in S_{imp}-S_{add}\\
0, & y_{i}\notin S_{imp}-S_{add}
\end{array} \right.
\end{equation}
where $y_i$ denote the label of inquiry class i. As for the symptom classification layer ouput $z$, the cross entropy loss of concurrent softmax is presented as:
\begin{equation}
    \label{consoftmax}
    \begin{split}
	\mathcal{L}_{sym}(y,z)&=-\sum_{i=1}^{C_{inq}}y_{i}\textrm{log}\sigma_{i}^{*} \\
	\textrm{with}\ \sigma_{i}^{*}&=\frac{e^{z_{i}}}{\sum_{j=1}^{C_{inq}}(1-y_{j})e^{z_{j}}+e^{z_{i}}}
\end{split}
\end{equation}
 where $C_{inq}$ denote the number of inquiry type. To balance label learning in $[S]$ sequence, the loss of a single $[S]$ is divided by the number of synchronous labels. With the synchronous learning, the training objective Eq.(\ref{pred_target}) transfer to
 \begin{equation}
    \label{syn}
 \prod_{i=1}^{|T_{imp}|}\prod_{j=i}^{|T_{imp}|}P(T_{imp}^{j}\mid S_{exp}, T_{imp}^{<i}) 
 \end{equation}
 , which is more approximate to the Eq.(\ref{ground_target}). Moreover, the synchronous learning helps improve training efficiency.

\paragraph{Repeated Sequence} Due to the autoregressive generation, the model is plagued with learning a specific order generation of implicit symptoms in each training step. This results in a training imbalance of symptoms generations in different orders. For example in Figure~\ref{overview}a, the model can learn to inquiry symptoms sequentially as $Sym_3\rightarrow Sym_4\rightarrow Sym_5$, but fail to learn to inquiry symptoms as like $Sym_5\rightarrow Sym_4\rightarrow Sym_3$, because of the unidirectional sequence generation.  To reduce the impact of unidirectional generation, we constructed the repeated sequences concatenated with the input sequence to learn different orders of sequence generation. As shown in Figure~\ref{orderless}, $[Sym_5^{'},[S]_2^{'},Sym_4^{'},[S]_3^{'}]$ sequence is added to enable that generate implicit symptoms as $Sym_5\rightarrow Sym_4\rightarrow Sym_3$. The repeated sequence consist of last (n-1) implicit symptoms, in a new order sequence randomly shuffled like sequence shuffle, and (n-1) $[S]$ tokens, set to predict symptoms in the new order. The repeated sequences share the first $[S]$ and \emph{END} symbol prediction with original sequence. Benefit from the ability of long dependency and the parallel computing in Transformer, we can expand input sequence with the repeated sequences to alleviate the bias between Eq.(\ref{pred_target}) and Eq.(\ref{ground_target}), and improve the training efficiency of orderless generation. Since the symptoms sequence is relatively short, we set the number of repeated sequences as 4.

\begin{figure}[h]
  \centering
  \includegraphics[width=0.85\columnwidth]{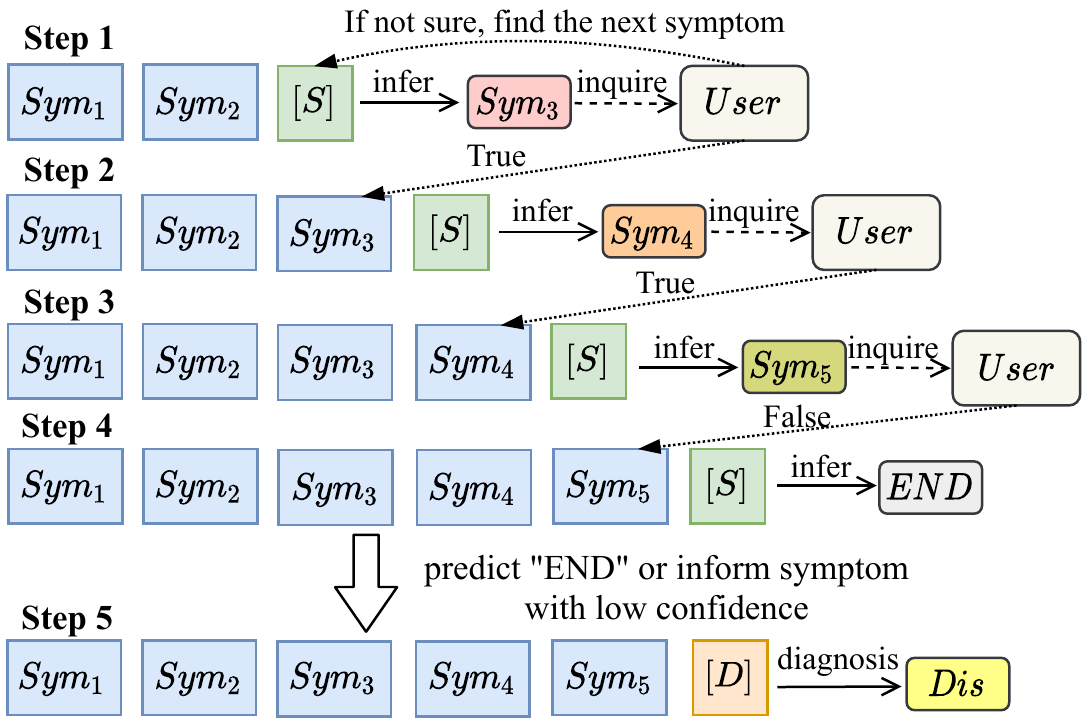}
  \caption{\label{inference}Schematic of inference process. \emph{User} is the user simulator.  }
\end{figure}

\subsection{Generative Inference}
During inference stage, we firstly inserted a symptom prediction token $[S]$ behind the symptoms sequence to calculate the probability distribution of symptoms. 
Then we mask the symptoms have been inquired and inquire the rest highest probability symptom. Next, the user simulator would determine if the inquired symptom is in the implicit symptoms set. If the symptom is not an implicit symptom, the model will find the next highest probability symptom to inquire user. Once the inquiry symptom is a implicit symptom, the user simulator will reply True or False for the symptom, and then the model insert it into the symptoms sequence and predict new probabilities of inquiry symptom, detailed as Figure~\ref{inference}. When the model predict \emph{END} symbol with probability greater than $\rho_e$ or infer the inquiry symptom with probability less than $\rho_p$, the model will stop symptom inquiry to diagnose disease. In disease diagnosis, we insert the disease prediction token $[D]$ into the symptoms sequence to predict the disease.

\section{Experiments}
\subsection{Datasets}
We evaluate our model on three public automatic diagnosis datasets, namely MuZhi dataset~\cite{wei2018task}, Dxy dataset~\cite{xu2019end} and Synthetic dataset~\cite{liao2020task}. MuZhi dataset and Dxy dataset are real-data from self-reports and the conversations. Synthetic dataset is a much bigger synthetic data constructed by symptom-disease dataset. The datasets statistics are shown in Table~\ref{dataset}. 

\begin{table}
  \centering
  \begin{tabular}{@{}lcccccc@{}}
    \toprule
    \small{\textbf{Dataset}} & \small{\textbf{\# Disease}} & \small{\textbf{\# Symptom}} & \small{\textbf{\# Training}} & \small{\textbf{\# Test}} \\
    \midrule
    MuZhi & 4 & 66 & 568 & 142 \\
    Dxy & 5 & 41 & 423 &104 \\
    Synthetic & 90 & 266 & 24,000 & 6,000 \\
    \bottomrule
  \end{tabular}
  \caption{Statistics of the three datasets.}
  \label{dataset}
\end{table}

\subsection{Experimental Details}
For all model setting, the train set and test set both use the original format, as shown in Table~\ref{dataset}. All the experiment is carried by 5 times and the final result is the average of the best results on test set. Diaformer and its variants use small transformer networks ($L$=5, $H$=512, $A$=6). For training, the learning rate is $5e-5$ and the batch size is 16. For inference, we set $\rho_e$ as 0.9 and set $\rho_p$ as 0.009 for MuZhi dataset, 0.012 for Dxy dataset and 0.01 for synthetic dataset.

\subsection{Comparison}
\paragraph{Baselines} Firstly, we use SVM to classify disease based on explicit symptoms without any symptom inquiry and name it SVM-exp to give a a minimum baseline of diagnosis accuracy. Then we have selected five competitive RL-based models as comparison, including Flat-DQN~\cite{wei2018task}, HRL~\cite{liao2020task}, KR-DS~\cite{xu2019end}, GAMP~\cite{xia2020generative} and PPO~\cite{Teixeira2021TheIO}. Besides, we add two SG-based models, namely Diaformer$_{\textrm{GPT2}}$ and Diaformer$_{\textrm{UniLM}}$, serve as strong SG-based baseline of Diaformer. Diaformer$_{\textrm{GPT2}}$ and Diaformer$_{\textrm{UniLM}}$ base on our symptom attention framework and train on the objective in GPT2~\cite{radford2019language} and UniLM~\cite{dong2019unified}, which are two classic sequence generation model fitted to automatic diagnosis. For fair comparison, Diaformer$_{\textrm{GPT2}}$ and Diaformer$_{\textrm{UniLM}}$ are extra added with the sequence shuffle training mechanisms, and use the same hyper parameters of Diaformer.

\begin{table*}
  \centering 
   \begin{tabular}{@{}lcccccccccccc@{}}
\toprule
\multirow{2}{*}{Model} & \multicolumn{4}{c}{MuZhi dataset}                             & \multicolumn{4}{c}{Dxy dataset}                              & \multicolumn{4}{c}{Synthetic dataset}                          \\ \cmidrule(l){2-5} \cmidrule(l){6-9} \cmidrule(l){10-13}  
                       & DAcc           & SRec           & ATurn         & Ttime       & DAcc           & SRec           & ATurn        & Ttime       & DAcc           & SRec           & ATurn         & Ttime        \\ \midrule
SVM-exp               & 0.673          & -          & -           & -         & 0.640          & -          & -          & -        & 0.341          & -           & -           & -           \\
Flat-DQN               & 0.690          & 0.301          & \textbf{3.1}           & 82m         & 0.720          & 0.322          & 2.9          & 141m        & 0.356          & 0.02           & \textbf{2.0}           & 50m          \\
HRL                    & 0.694          & 0.276          & 3.5           & 162m        & 0.695          & 0.161          & \textbf{2.4}          & 35m         & 0.496          & 0.338          & 8.4           & 673m         \\
KR-DS$\ddagger$                  & 0.730          & -              & 3.4          & -           & 0.740          & -              & -            & -           & -              & -              & -             & -            \\
GAMP$\ddagger$                   & 0.730          & -              & -             & -           & 0.769          & -              & 2.7         & -           & -              & -              & -             & -            \\
PPO$\ddagger$                    & 0.732          & -              & 6.3           & -           & 0.746          & -              & 3.3         & -           & 0.618          & -              & 12.6         & -            \\ \midrule
Diaformer$_{\textrm{GPT2}}$          & 0.740          & 0.745          & 15.3          & \textbf{2m}          & 0.811          & 0.798          & 11.2          & \textbf{2m}          & 0.724          & 0.890          & 12.9          & 19m          \\
Diaformer$_{\textrm{UniLM}}$        & 0.739          & 0.742          & 15.2          & \textbf{2m}          & 0.817          & 0.817          & 11.2          & 3m          & 0.722          & 0.886          & 12.7          & 33m          \\
Diaformer              & \textbf{0.742} & \textbf{0.752} & 15.3 & \textbf{2m} & \textbf{0.829} & \textbf{0.827} & 13.1 & \textbf{2m} & \textbf{0.733} & \textbf{0.906} & 13.7 & \textbf{17m} \\ \bottomrule
\end{tabular}
\caption{Results on three datasets. DAcc is the accuracy of diagnosis; SRec is the recall of the implicit symptoms; ATurn is the average of symptom inquiry turn; Ttime indicates the training time to get the best diagnosis result running on a 1080Ti GPU; Ttime's unit ``m'' indicate minute; $\ddagger$ marks the results reported by the original papers.}
  \label{sympre}
\end{table*}

\paragraph{Overall Performance} According to the task definition in \cite{wei2018task} and \cite{liao2020task}, we set the maximum inquiry turn as 20. We evaluate all the model by three metrics, which are diagnosis accuracy, average inquiry turns and recall of the implicit symptoms. The recall of the implicit symptoms is a significant metric for SG-based models, which aim to inquire the implicit symptoms out as much as possible and then diagnose the disease. Besides, we add a training time metric to evaluate the training efficiency of models. The results on three datasets are shown in Table~\ref{sympre}. In the results, Diaformer overwhelmingly outperforms other models in diagnosis accuracy with highest training efficiency. Especially on Dxy dataset and Synthetic dataset, our model outperforms the current state-of-the-art model by 6\% and 11.5\% in diagnosis accuracy. Note that SG-based models have considerably high recall of implicit symptoms and perform much better on bigger dataset, i.e. Synthetic dataset. The upsurge on bigger dataset demonstrates the high training efficiency of SG-based models, which only need to be trained in 2 minutes on Dxy dataset and MuZhi dataset. It indicates that the symptoms sequence generation has considerable potential in automatic diagnosis. Additionally, Diaformer surpass the other SG-based mdoel, which demonstrates the symptom attention framework and the orderless mechanisms can further improve the performance and training efficiency in SG-based models. Moreover, we observe that SG-based models request more inquiry turns, due to the higher recall of symptoms lead to more inquiry turns. For simulate the process of doctor diagnosis, SG-based models tend to generate more implicit symptoms. For this limitation, we conduct the experiments of smaller limited turns.

\begin{table*}[t]
  \centering
   \begin{tabular}{llccccccccc}
\toprule
\multirow{2}{*}{\begin{tabular}[c]{@{}l@{}}Max\\ turn\end{tabular}} & \multirow{2}{*}{Model} & \multicolumn{3}{c}{MuZhi dataset}              & \multicolumn{3}{c}{Dxy dataset}                & \multicolumn{3}{c}{Synthetic dataset}        \\ 
\cmidrule(l){3-5} \cmidrule(l){6-8} \cmidrule(l){9-11}  
                                                                    &                        & DAcc           & SRec           & ATurn        & DAcc           & SRec           & ATurn        & DAcc           & SRec           & ATurn      \\ \midrule
\multirow{5}{*}{5}                                                  & Flat-DQN               & 0.641     & 0.292     & 2.9       & 0.647     & 0.311    & 2.5      & 0.356       & 0.02       & \textbf{2.0}          \\
                                                                    & HRL                    & 0.676     & 0.265     & \textbf{2.8}       & 0.702     & 0.152    & \textbf{1.9}      & 0.443       & 0.24       & 4.3        \\
                                                                    & Diaformer$_{\textrm{GPT2}}$          & 0.721     & 0.463     & 5.0       & 0.743     & 0.543    & 4.9     & 0.492       & 0.455       & 5.0       \\
                                                                    & Diaformer$_{\textrm{UniLM}}$        & \textbf{0.722}     & 0.466     & 5.0       & 0.759     & 0.539    & 4.9     & 0.492       & 0.454       & 5.0       \\
                                                                    & Diaformer              & \textbf{0.722}     & \textbf{0.472}     & 5.0       & \textbf{0.767}     & \textbf{0.545}    & 4.8     & \textbf{0.494}       & \textbf{0.461}       & 4.9       \\\midrule
\multirow{5}{*}{10}                                                 & Flat-DQN               & 0.683     & 0.296     & \textbf{3.0}       & 0.715     & 0.322    & 2.7      & 0.356       & 0.02       & \textbf{2.0}          \\
                                                                    & HRL                    & 0.697     & 0.266     & 3.3       & 0.718     & 0.159    & \textbf{2.3}      & 0.488       & 0.307      & 7.4        \\
                                                                    & Diaformer$_{\textrm{GPT2}}$          & 0.728     & 0.646     & 10.0       & 0.794     & 0.738    & 9.2      & 0.627       & 0.726      & 9.5        \\
                                                                    & Diaformer$_{\textrm{UniLM}}$        & \textbf{0.732}     & 0.652     & 9.8       & 0.804     & 0.758    & 9.0      & 0.630       & 0.723      & 9.4        \\
                                                                    & Diaformer              & 0.731     & \textbf{0.655}     & 9.8       & \textbf{0.806}     & \textbf{0.778}    & 9.6      & \textbf{0.632}       & \textbf{0.736}      & 9.6        \\\midrule
\multirow{5}{*}{15}                                                 & Flat-DQN               & 0.683     & 0.297     & \textbf{3.0}         & 0.712     & 0.32     & 2.7      & 0.356       & 0.02       & \textbf{2.0}          \\
                                                                    & HRL                    & 0.702     & 0.272     & 3.4       & 0.718     & 0.159    & \textbf{2.3}      & 0.499       & 0.322      & 8.3        \\
                                                                    & Diaformer$_{\textrm{GPT2}}$          & 0.733     & 0.724     & 13.7      & 0.807     & 0.793     & 10.6      & 0.704        & 0.849      & 12.0       \\
                                                                    & Diaformer$_{\textrm{UniLM}}$        & 0.733     & 0.717     & 13.6      & 0.814     & 0.804     & 11.1      & 0.702        & 0.847      & 11.9       \\
                                                                    & Diaformer               & \textbf{0.742} & \textbf{0.731} & 13.8 & \textbf{0.828} & \textbf{0.826} & 12.4 & \textbf{0.711} & \textbf{0.866} & 12.6 \\
\bottomrule
\end{tabular}
\caption{Results with smaller different limited turns. DAcc, SRec and ATurn are same as Table~\ref{sympre}.}
  \label{DAcc}
\end{table*}

\paragraph{Diagnosis with Smaller Limited Turns} Considering higher recall of implicit symptoms requesting more inquiry turns that is unfair to the less inquiry turns model, we conduct the experiments with smaller maximum turns, including 5, 10, 15. Due to the limitation of not being open source for some models, we conduct this experiment on two RL-based model and all SG-based models. Table~\ref{DAcc} shows the result of disease diagnosis with smaller limited turns. Overall, Diaformer still outperform other models in the limit of smaller turns in terms of recall of implicit symptoms and diagnosis accuracy. Note that in the limit of smaller turns, Diaformer always recall more implicit symptoms than the other SG-based models with almost equal turns, which indicates the higher performance of Diaformer. On the Synthetic dataset, SG-based models outperform the PPO model with smaller average turns. In the limit of 5 turns, Diaformer obtains competitive results of diagnosis accuracy with comparably few inquiry turns. The results of smaller limited turn indicate that SG-based models can still perform automatic diagnosis well with the smaller inquiry turns.

\begin{table*}[t]
  \centering
   \begin{tabular}{@{}llccccccccc@{}}
\toprule
\multirow{2}{*}{\#} & \multirow{2}{*}{Model} & \multicolumn{3}{c}{MuZhi dataset}              & \multicolumn{3}{c}{Dxy dataset}                & \multicolumn{3}{c}{Synthetic dataset}           \\ \cmidrule(l){3-5} \cmidrule(l){6-8} \cmidrule(l){9-11} 
                    &                        & DAcc           & SRec          & ATurn         & DAcc           & SRec           & ATurn        & DAcc           & SRec           & ATurn         \\ \midrule
1                   & Diaformer              & \textbf{0.742} & \textbf{0.752} & 15.3          & \textbf{0.829} & \textbf{0.827} & 13.1          & \textbf{0.733} & \textbf{0.906} & 13.7          \\
2                   & w/o Sequence Shuffle                   & 0.737          & 0.723         & 17.3          & 0.825          & 0.824          & 14.4 & 0.658          & 0.730          & 13.2 \\
3                   & w/o Synchronous Learning                   & \textbf{0.742} & 0.738         & 14.3          & 0.826          & 0.790          & 11.1          & 0.725          & 0.891          & 12.8          \\
4                   & w/o Repeated Sequence                   & 0.735          & 0.705         & \textbf{13.4} & 0.817          & 0.773          & \textbf{11.0}          & 0.713          & 0.877          & \textbf{12.5}          \\ \bottomrule
\end{tabular}
\caption{Results of ablation study. DAcc, SRec and ATurn are same as Table~\ref{sympre}.}
  \label{ablation}
\end{table*}

\subsection{Ablation Study}
\begin{figure}[t]
\centering
{ \label{fig:b}
\includegraphics[width=0.48\columnwidth]{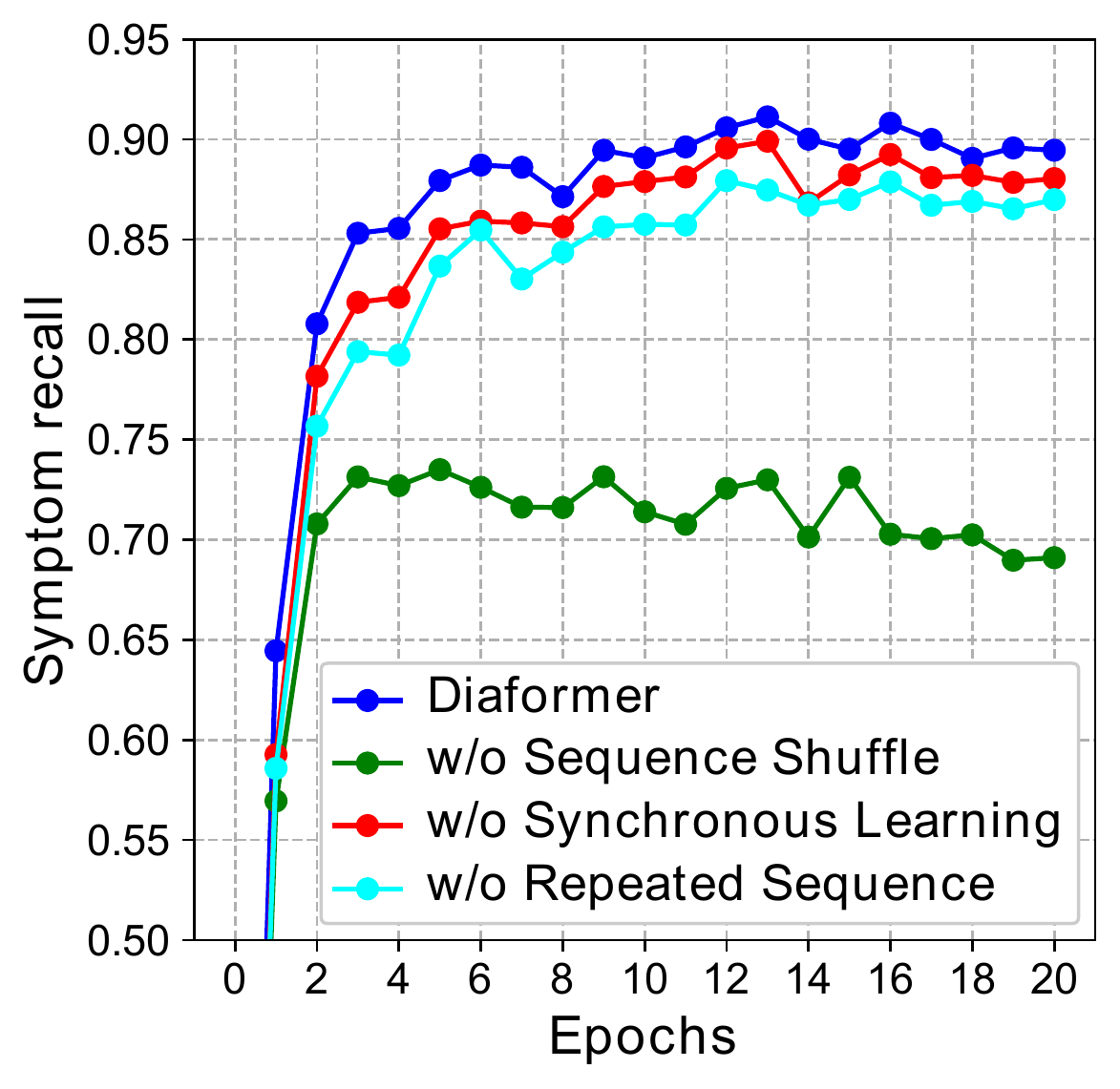}
}
{ \label{fig:a}
\includegraphics[width=0.48\columnwidth]{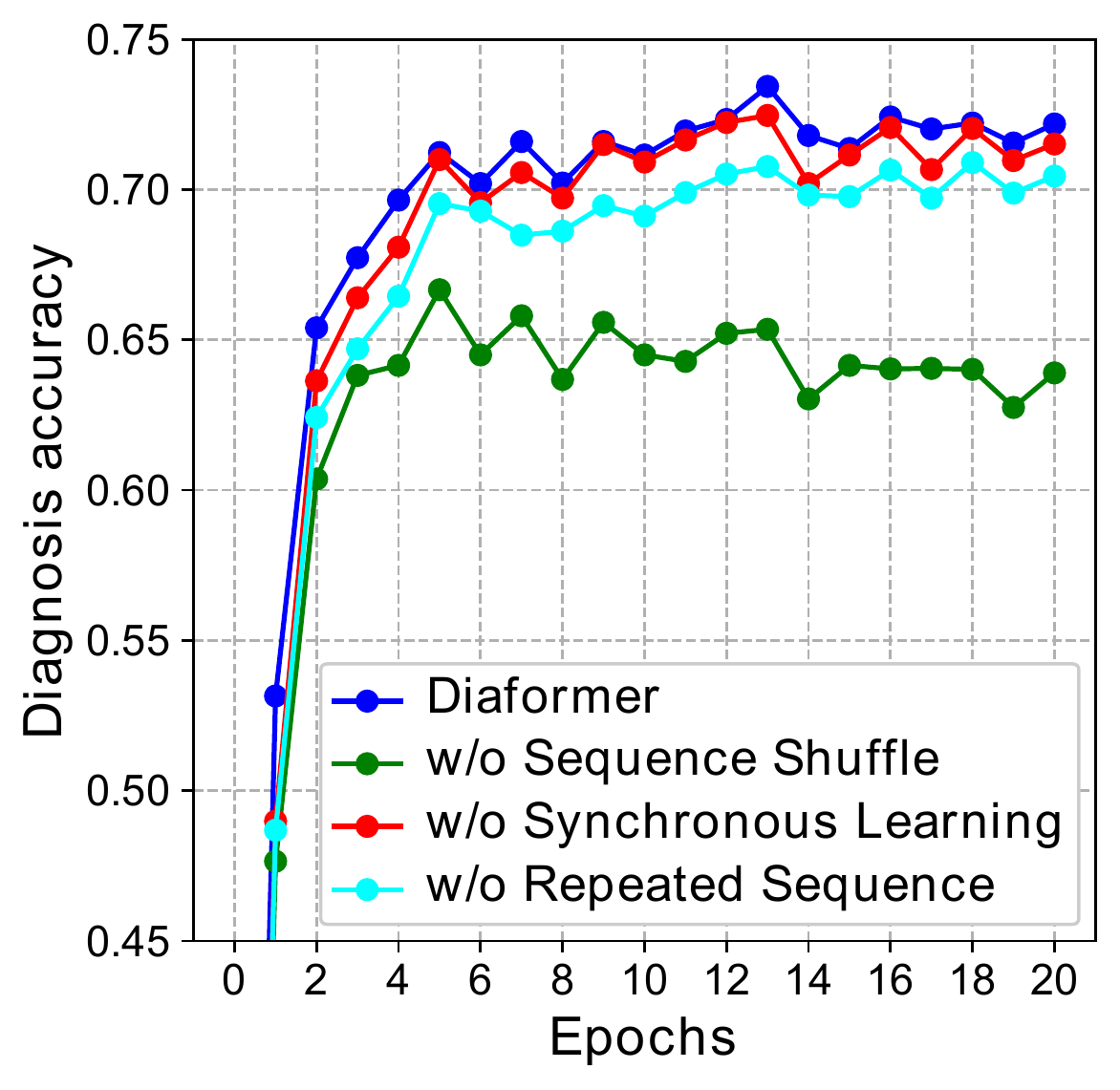}
}
\label{figAblation}
\caption{\label{ablation-fig}Results of ablation studies on the synthetic dataset.}
\label{fig}
\end{figure}

We perform an ablation study to understand the importance of three training mechanisms of orderless generation. In Table~\ref{ablation}, we compare three Diaformer variants (rows 2 - 4) without one of the three orderless training mechanisms. As shown in Table~\ref{ablation}, we can see each mechanism contribute to improve the performance. Without sequence shuffle, the model obtain lower recall of symptoms with higher inquiry turns. Without synchronous learning or repeated sequence, the accuracy of diagnosis and the recall of symptoms both reduce. Specifically, Figure~\ref{ablation-fig} show the results of on Synthetic dataset in the series of training epoch as the same parameter initialization, in which Diaformer obviously performs best on diagnosis accuracy and recall of implicit symptoms. It indicates that orderless training mechanism help to improve the performance of symptoms sequence generation in automatic diagnosis.

\section{Related Work}
\label{related_work}

\paragraph{Automatic Diagnosis} There are some previous works for automatic diagnosis, which mostly use RL~\cite{tang2016inquire,kao2018context,peng2018refuel,wei2018task,xu2019end,liao2020task,Hou2021ImperfectAD,Teixeira2021TheIO}. \citet{tang2016inquire} propose neural symptom checking, which adopts reinforcement learning to simultaneously conduct symptom inquiries and diagnose. Based on the work of \cite{tang2016inquire}, \citet{kao2018context} employ hierarchical reinforcement learning to make a joint diagnostic decision and introduce context to make the symptom checker context aware. \citet{wei2018task} use a Deep Q-network from conversation with patients to collect additional symptoms, which can greatly improve the accuracy of diagnosis. \citet{xu2019end} introduce prior medical knowledge to guide policy learning. \citet{liao2020task} classify diseases into several groups and uses a hierarchy of two levels for automatic disease diagnosis using HRL methods. \citet{xia2020generative} propose a policy gradient framework based on the Generative Adversarial Network to optimize the RL model. Recently, \citet{Hou2021ImperfectAD} propose a multi-level reward RL-based model and \citet{Teixeira2021TheIO} customize the settings of the reinforcement learning leveraging the dialogue data. 

\paragraph{Sequence Generation} Sequence generation task aims to generate a target sequence condition on a source input. It covers many areas with a lot of tasks~\cite{zhang2020multi}. Among them, the natural language generation (NLG) have achieved great success with the development of neural networks. Recently, the Transformer-based models have obtained superior performance in NLG, such as \cite{radford2019language,yang2019xlnet,DBLP:journals/corr/abs-1905-02450,brown2020language,bao2020unilmv2,xiao2020ernie,Sun2021ERNIE3L}. Most of them train as auto-regressive (AR), in which the probability of an output token depends on all previous tokens. Based on AR training objective, the sequence generation model can learn all the target tokens in parallel. Besides, some of them use the additional artificial symbol sequence~\cite{xiao2020ernie,brown2020language} or combine Masked Language model~\cite{devlin2018bert} as auto-encoding (AE)~\cite{dong2019unified,bao2020unilmv2} to enhance the model. With more and more relevant models being proposed, Transformer has shown great potential in sequence generation.

\section{Conclusion}
In this work, we reformulate the automatic diagnosis problem as a sequence generation task and propose a symptom attention framework for automatic diagnosis with symptoms sequence generation. Besides, we propose three orderless training mechanisms to alleviate the bias of the discrepancy between the sequential generation and the disorder of symptoms. Experimental results show that our model outperforms other models on three datasets of automatic diagnosis and demonstrates the potential of symptoms sequence generation in automatic diagnosis. Future work includes incorporating Diaformer into task-oriented dialogue system of diagnosis and effectively lessen the inquiry turns.

\section*{Acknowledgements}
We thank all the reviewers for their constructive comments and useful suggestions. This work is supported by the Natural Science Foundation of China (Grant No. 61872113), Strategic Emerging Industry Development Special Funds of Shenzhen (No. XMHT20190108009), and Fundamental Research Fund of Shenzhen (No. JCYJ20190806112210067).

\section*{Ethics Statement}
The problem of facilitating diagnosis is important in artiﬁcial intelligence applications. Our method shows promising accuracy and noticeable efficiency on automatic diagnosis, which is a multi-step reasoning problem as well as a symptom checking task in \cite{tang2016inquire}. Besides strong performance, the automatic diagnosis models learned from the insufficient and incomplete dataset has considerable risk of predicting error that may cause seriously harm. Under the ethical considerations, we suggest users regard it as an auxiliary tool that can help doctors make diagnose or help to give patients some advice.

In fact, the proposed Diaformer is not limited to the automatic diagnosis problem. It can extend to use in some decision-making problems or RL problems through slight change. As for the usage on other problems, we suggest users design more intermediate-state tokens along with the decision tokens to form a decisions sequence and adjust the type embedding and attention mask mechanism for specific problems. Different from Decision Transformer~\cite{add1} in the typical RL problems, our model tends to learn the relationship among the decisions directly and focus on alleviating the order bias for orderless or non-sequential RL problems. Note that all the decision-making models run the risk of biased prediction in real-life application scenarios, and be careful to use them.

\bibliography{aaai22}

\clearpage
\section*{Appendices}
\subsection{A\quad Why Use Concurrent Softmax}
In the synchronous learning, we attend to train the output of $[S]$ to predict multiple symptoms, which can be regarded as a multi-label task. Normally, the Bernoulli parameter for each symptom inquiry label is considered as a good choice to the multi-label task. However, there are three reason that we choose the concurrent softmax: (1) the synchronous labels are sparse, due to the number of implicit symptoms is much less than the number of symptom inquiry types. Training as Bernoulli parameter will perform inefficiently. (2) The synchronous learning is a training mechanism in the training stage, while we still use softmax to compute the probability distribution of symptom inquiry. As a result, if we use the Bernoulli parameter, it will lead to a big discrepancy between training and inference. In contrast, the concurrent softmax is almost same as softmax, used to learn the probability distribution of symptom inquiry. (3) Due to the imbalance of the symptom inquiry label, the independent learning in the Bernoulli parameter give rise to a serious overfitting problem. In our pilot experiments, the concurrent softmax indeed perform much better than the Bernoulli parameter.

\begin{table*}
  \centering
  \begin{tabular}{@{}lcccccc@{}}
    \toprule
    \textbf{Dataset} & \textbf{\# Disease} & \textbf{\# Symptom} & \textbf{\# ave ex. sym.} &\textbf{\# ave im. sym.} & \textbf{\# Training} & \textbf{\# Test} \\
    \midrule
    MuZhi dataset & 4 & 66 & 2.35 & 3.26 & 568 & 142 \\
    Dxy dataset & 5 & 41 & 3.07 & 1.67 & 423 &104 \\
    Synthetic dataset & 90 & 266 & 1 & 2.6 & 24,000 & 6,000 \\
    \bottomrule
  \end{tabular}
  \caption{Statistics of the three datasets. Ave ex. sym. and ave im. sym. are the average number of explicit and implicit symptoms.}
  \label{dataset2}
\end{table*}

\subsection{B\quad Dataset Details}
\paragraph{MuZhi Dataset} The MuZhi dataset is constructed by \cite{wei2018task}. The data is collected from the pediatric department of a Chinese online medical website, which is a popular website for users to consult doctors online. Usually, a patient will provide a self-report to indicate his or her basic information.Then the doctor will initiate a conversation to collect more information and make a diagnosis based on parent's self-report and conversations. The doctor can obtain additional symptoms not mentioned in the self-report during the conversation. For each patient, the doctor will give the final diagnosis as the label. In the dataset, symptoms extracted from self-report are regarded as explicit symptoms and the ones extracted from conversation are implicit symptoms. More detailed dataset statistics are shown in Table \ref{dataset2}.

\paragraph{Dxy Dataset} The Dxy dataset is constructed by \cite{xu2019end} and the data is collected from a Chinese online  health-care community, where users asking doctors for medical diagnosis or professional medical advice. The dataset contains five types of diseases, including allergic rhinitis, upper respiratory infection, pneumonia, children hand-foot-mouth disease, and pediatric diarrhea. And \cite{xu2019end} extract the symptoms that appear in self-reports and conversation and normalize them into 41 symptoms. More detailed dataset statistics are shown in Table \ref{dataset2}. Similar to the MuZhi dataset, symptoms appearing in self-reports are regarded as symptoms while the others are implicit symptoms.

\paragraph{Synthetic Dataset} The Synthetic dataset is constructed by \cite{liao2020task} following \cite{kao2018context}, which is based on symptom-disease database called SymCat. There are 801 diseases in the database and \cite{liao2020task} classify them into 21 departments (groups) according to International Classification of Diseases (ICD-10-CM) and choose 9 representative departments from the database. Each department contains top 10 diseases according to the occurrence rate in the Centers for Disease Control and Pre-vention (CDC) database. More detailed dataset statistics are shown in Table \ref{dataset2}.

\subsection{C\quad Comparison Models}
\paragraph{Minimum Baseline}
\begin{itemize}
    \item \textbf{SVM-exp}: To indicate the significance of symptom inquiry and give the minimum boundaries in automatic diagnosis, we use SVM model learned to distinguish the disease on explicit symptoms directly without any symptom inquiry. The SVM we used is based on the linear kernel and adopt one-vs-one scheme to support multiclass.
\end{itemize}

\paragraph{RL-based Models}
\begin{itemize}
    \item \textbf{Flat-DQN}: This is the model from \cite{wei2018task}, which has one layer policy and an action space including both symptoms and diseases.
    \item \textbf{HRL}: This is the model from \cite{liao2020task}, which integrates a hierarchical policy of two levels into the dialogue policy learning. The high level policy consists of a model named master that is responsible for triggering a model in low level, and the low level policy consists of several symptom checkers and a disease classifier.
    \item \textbf{KR-DS}: This is the model from \cite{xu2019end}, which seamlessly incorporates rich medical knowledge graph into the topic transition in dialogue management. The model use a novel Knowledge-routed Deep Q-network (KR-DQN) ,which integrates a relational refinement branch for encoding relations among different symptoms and symptom-disease pairs, and a knowledge-routed graph branch for topic decision-making.
    \item \textbf{GAMP}: This is the model from \cite{xia2020generative}. GAMP is a generative adversarial regularized Mutual information Policy gradient framework (GAMP) for automatic diagnosis, that based on the Generative Adversarial Network (GAN) to optimize the RL model for automatic diagnosis.
    \item \textbf{PPO}: This is the model from \cite{Teixeira2021TheIO}, which is proposed to leverage the models learned from the dialogue data to customize the settings of the reinforcement learning for more efficient action space exploration.
\end{itemize}

\paragraph{SG-based variant models}
\begin{itemize}
    \item \textbf{Diaformer$_{GPT2}$}: This is a Diaformer variant model. Diaformer$_{GPT2}$ uses the symptom attention framework and ${GPT2}$~\cite{radford2019language} SG training objective, that each symptom token, including explicit symptoms, predict the next symptom in the symptoms sequence without $[S]$ token. To be fair comparison, we add the sequence shuffle mechanism for it. GPT2 is a classics autoregression generation model with high efficiency of training. The Diaformer$_{GPT2}$ is compared as a conventional generative model.
    \item \textbf{Diaformer$_{UniLM}$}: This is a Diaformer variant model based on the symptom attention framework using the UniLM~\cite{dong2019unified} training objective. Basd on the UniLM training setting, 1/3 of the time we use the bidirectional objective, 1/3 of the time we employ the sequence-to-sequence generation objective, and the rest 1/3 time we use unidirectional generation objective, which replace left-to-right and right-to-left in UniLM since sequence shuffle. For the bidirectional objective, the model allows all tokens to attend to each other in prediction and mask a symptom token at each time. As for the mask, 80\% of the time we replace the token with [MASK], 10\% of the time with a random token, and keeping the original token for the rest. For the sequence-to-sequence generation objective, we asked the model to generate implicit symptoms sequence by explicit symotoms. We set the UniLM variant comparison to test the advanced SG model with auto-encoder on automatic diagnosis.
\end{itemize}

\subsection{D\quad Model Implementation Details}
Because it is small scale of automatic diagnosis datasets, Diaformer use a small transformer network ($L$=5, $H$=512, $A$=6), where $L$ denotes the number of layers, $H$ denotes the hidden size, $A$ denotes the number of attention heads. The learning rate is set to $5e-5$ and the batch size is 16. For the inference step, we set $\rho_e$=0.9 and set $\rho_p$ as 0.009 for MuZhi dataset, 0.012 for Dxy dataset and 0.01 for synthetic dataset. The hyper parameters of Diaformer$_{GPT2}$ and Diaformer$_{UniLM}$ are the same as Diaformer.

\subsection{E\quad Standard Errors of Result}
Due to space limitation, we give the standard error of our models in each experiment, which is carried by 5 times. The standard errors are given in Table~\ref{error}.

\begin{table*}[t]
  \centering
  \begin{tabular}{llccccccccc}
\toprule
\multirow{2}{*}{\begin{tabular}[c]{@{}l@{}}Max\\ turn\end{tabular}} & \multirow{2}{*}{Model} & \multicolumn{2}{c}{MuZhi dataset}              & \multicolumn{2}{c}{Dxy dataset}                & \multicolumn{2}{c}{Synthetic dataset}        \\ 
\cmidrule(l){3-4} \cmidrule(l){5-6} \cmidrule(l){7-8}  
                                                                    &                        & DAcc           & SRec                   & DAcc           & SRec                   & DAcc           & SRec            \\ \midrule
\multirow{3}{*}{5}
                                                                    & Diaformer$_{GPT2}$          & 0.721\footnotesize{$\pm.008$}     & 0.463\footnotesize{$\pm.010$}            & 0.743\footnotesize{$\pm.004$}     & 0.543\footnotesize{$\pm.017$}         & 0.492\footnotesize{$\pm.004$}       & 0.455\footnotesize{$\pm.004$}              \\
                                                                    & Diaformer$_{UniLM}$        & 0.722\footnotesize{$\pm.005$}     & 0.466\footnotesize{$\pm.011$}            & 0.759\footnotesize{$\pm.020$}     & 0.539\footnotesize{$\pm.011$}         & 0.492\footnotesize{$\pm.004$}       & 0.454\footnotesize{$\pm.004$}              \\
                                                                    & Diaformer              & 0.722\footnotesize{$\pm.003$}     & 0.472\footnotesize{$\pm.017$}            & 0.767\footnotesize{$\pm.007$}     & 0.545\footnotesize{$\pm.031$}         & 0.494\footnotesize{$\pm.003$}       & 0.461\footnotesize{$\pm.002$}              \\\midrule
\multirow{3}{*}{10}                                                 
                                                                    & Diaformer$_{GPT2}$          & 0.728\footnotesize{$\pm.008$}     & 0.646\footnotesize{$\pm.010$}            & 0.794\footnotesize{$\pm.009$}     & 0.738\footnotesize{$\pm.015$}          & 0.627\footnotesize{$\pm.005$}       & 0.726\footnotesize{$\pm.004$}              \\
                                                                    & Diaformer$_{UniLM}$        & 0.732\footnotesize{$\pm.004$}     & 0.652\footnotesize{$\pm.013$}            & 0.804\footnotesize{$\pm.004$}     & 0.758\footnotesize{$\pm.007$}          & 0.630\footnotesize{$\pm.001$}       & 0.723\footnotesize{$\pm.001$}              \\
                                                                    & Diaformer              & 0.731\footnotesize{$\pm.008$}     & 0.655\footnotesize{$\pm.003$}            & 0.806\footnotesize{$\pm.003$}     & 0.778\footnotesize{$\pm.021$}          & 0.632\footnotesize{$\pm.003$}       & 0.736\footnotesize{$\pm.002$}              \\\midrule
\multirow{3}{*}{15}                             
                                                                    & Diaformer$_{GPT2}$          & 0.733\footnotesize{$\pm.007$}     & 0.724\footnotesize{$\pm.004$}           & 0.807\footnotesize{$\pm.008$}     & 0.793\footnotesize{$\pm.021$}           & 0.704\footnotesize{$\pm.001$}        & 0.849\footnotesize{$\pm.002$}             \\
                                                                    & Diaformer$_{UniLM}$        & 0.733\footnotesize{$\pm.005$}     & 0.717\footnotesize{$\pm.011$}           & 0.814\footnotesize{$\pm.013$}     & 0.804\footnotesize{$\pm.024$}           & 0.702\footnotesize{$\pm.001$}        & 0.847\footnotesize{$\pm.003$}             \\
                                                                    & Diaformer               & 0.742\footnotesize{$\pm.009$} & 0.731\footnotesize{$\pm.014$}  & 0.828\footnotesize{$\pm.016$} & 0.826\footnotesize{$\pm.028$}  & 0.711\footnotesize{$\pm.002$} & 0.866\footnotesize{$\pm.004$}  \\\midrule
\multirow{3}{*}{20} & Diaformer$_{GPT2}$          & 0.740\footnotesize{$\pm.002$}          & 0.745\footnotesize{$\pm.007$}                  & 0.811\footnotesize{$\pm.010$}          & 0.798\footnotesize{$\pm.018$}                 & 0.724\footnotesize{$\pm.002$}          & 0.890\footnotesize{$\pm.001$}                        \\
& Diaformer$_{UniLM}$        & 0.739\footnotesize{$\pm.008$}          & 0.742\footnotesize{$\pm.011$}                         & 0.817\footnotesize{$\pm.013$}          & 0.817\footnotesize{$\pm.014$}                     & 0.722\footnotesize{$\pm.001$}          & 0.886\footnotesize{$\pm.002$}                       \\
& Diaformer             & 0.742\footnotesize{$\pm.013$} & 0.752\footnotesize{$\pm.009$}  & 0.829\footnotesize{$\pm.016$} & 0.827\footnotesize{$\pm.032$}   & 0.733\footnotesize{$\pm.002$} & 0.906\footnotesize{$\pm.003$}  \\ \midrule
\multicolumn{8}{l}{\emph{Ablation Study}} \\
\multicolumn{2}{l}{w/o Sequence Shuffle}        & 0.737\footnotesize{$\pm.005$}          & 0.723\footnotesize{$\pm.020$}                   & 0.825\footnotesize{$\pm.015$}          & 0.824\footnotesize{$\pm.028$}           & 0.658\footnotesize{$\pm.003$}          & 0.730\footnotesize{$\pm.006$}           \\
\multicolumn{2}{l}{w/o Synchronous Learning}                   & 0.742\footnotesize{$\pm.007$} & 0.738\footnotesize{$\pm.022$}                   & 0.826\footnotesize{$\pm.008$}          & 0.790\footnotesize{$\pm.024$}                    & 0.725\footnotesize{$\pm.002$}          & 0.891\footnotesize{$\pm.003$}                    \\
 \multicolumn{2}{l}{w/o Repeated Sequence}                   & 0.735\footnotesize{$\pm.005$}          & 0.705\footnotesize{$\pm.023$}         & 0.817\footnotesize{$\pm.013$}          & 0.773\footnotesize{$\pm.027$}                   & 0.713\footnotesize{$\pm.001$}           & 0.877\footnotesize{$\pm.003$}                     \\ 
\bottomrule
\end{tabular}
\caption{Results with the standard error of our models. DAcc and SRec  are the same as Table~\ref{sympre}. Note that 20 maximum turn is the original setting of the task.}
  \label{error}
\end{table*}

\subsection{F\quad Examples of Result}
To help readers to comprehend the models output on the datasets, we give a success example and a failure example on Synthetic dataset, that shown in Table~\ref{example}. In the task, models are asked to sequentially inquire the implicit symptoms based on explicit symptoms and then diagnose the disease by the explicit symptoms and the inquired implicit symptoms. Besides, in the outputs of Diaformer, we find that model tend to inquire more symptoms in the failure examples, while inquire less symptoms in success example, e.g. the examples in Table~\ref{example}. It indicates the model has learned the diagnosis paradigm to some extent.

\begin{table*}[t]
  \centering
  \begin{tabular}{l}
\toprule
\textbf{Examples on Synthetic dataset} \\ \midrule
\makecell*[l]{
\emph{Success example}
}\\
\makecell*[l]{
\textbf{explicit\_symptoms:} \\ \{\textcolor{red}{Heavy menstrual flow}: \emph{True}\} \\
\textbf{implicit\_symptoms (GOAL):} \\ \{\textcolor{blue}{Unpredictable menstruation}: \emph{True}, \textcolor{blue}{Involuntary urination}: \emph{True}\}\\
\textbf{disease\_tag (GOAL):} \\ Endometrial hyperplasia
\\
\textbf{Sequential inquiries (PRED):} \\ Vaginal bleeding after menopause$\rightarrow$\textcolor{blue}{Unpredictable menstruation}$\rightarrow$Vaginal discharge$\rightarrow$Pelvic pain$\rightarrow$\\\textcolor{blue}{Involuntary urination}$\rightarrow$\colorbox[rgb]{0.93,0.93,0.93}{\emph{END}} \\
\textbf{Disease Diagnosis (PRED):} \\Endometrial hyperplasia $\surd$ \\} \\ \midrule 
\makecell*[l]{
\emph{Failure example}
}\\
\makecell*[l]{
\textbf{explicit\_symptoms:} \\ \{\textcolor{red}{Peripheral edema}: \emph{True}\} \\
\textbf{implicit\_symptoms (GOAL):} \\ \{\textcolor{blue}{Shortness of breath}: \emph{True}, \textcolor{blue}{Weakness}: \emph{True}, \textcolor{blue}{Sharp abdominal pain}: \emph{True}, \textcolor{blue}{Sharp chest pain}: \emph{True}\}\\
\textbf{disease\_tag (GOAL):} \\ Acute kidney injury
\\
\textbf{Sequential inquiries (PRED):} \\ \textcolor{blue}{Shortness of breath}$\rightarrow$\textcolor{blue}{Weakness}$\rightarrow$Nausea$\rightarrow$Leg swelling$\rightarrow$Difficulty breathing$\rightarrow$\textcolor{blue}{Sharp abdominal pain}$\rightarrow$\textcolor{blue}{Sharp chest pain}\\$\rightarrow$Vomiting$\rightarrow$Fluid retention$\rightarrow$Dizziness$\rightarrow$Hand or finger swelling$\rightarrow$Cough$\rightarrow$\colorbox[rgb]{0.93,0.93,0.93}{\emph{END}} \\
\textbf{Disease Diagnosis (PRED):} \\Fluid overload $\times$ \\} \\
 \bottomrule
\end{tabular}
\caption{Diagnosis examples of Diaformer}
\label{example}
\end{table*}

\end{document}